# Physics-Guided Masked Multi-Task Network for Edge-Friendly Battery Health Diagnostics from Stochastically Fragmented Charging Profiles

Shuhao Chen[1], Tianyu Shi[1], Chengyi Tu[1*]

[1]School of Economics and Management, Zhejiang Sci-Tech University, Hangzhou, 310018, China.

*Corresponding author. Email: chengyitu1986@gmail.com

**Abstract**

Accurate State of Health (SOH) estimation is imperative for battery operational safety; nevertheless, translating data-driven prognostics to commercial Battery Management Systems (BMSs) remains restricted by pervasive data incompleteness, the latent trajectory of macroscopic SOH deterioration, and gradient conflict within multi-task frameworks. Addressing these structural challenges, this study introduces the Physics-Guided Masked Multi-Task Network (PG-M2TN), a highly compact architecture engineered to synthesize physical principles with deep representations. The proposed topology operates through three synergistic mechanisms. First, an optimized BiLSTM–Attention backbone captures temporal ageing dynamics with strict linear computational efficiency. Second, a Masked Autoencoder (MAE) reconstruction branch enforces the deterministic recovery of intact voltage–current profiles from stochastic observations. This generative pathway inherently serves as a structural regularizer, aligning the shared latent manifold to explicitly mitigate multi-task negative transfer. Third, a dual-stream prediction head fuses macroscopic SOH estimation with the Voltage Distortion Ratio (VDR), a microscopic leading indicator driven by polarization physics, thereby enabling the network to predictively anticipate late-life capacity collapse. Extensive empirical evaluations across five cross-chemistry lithium-ion datasets confirm the architectural superiority of PG-M2TN, yielding a global RMSE of 0.0781. The model preserves rigorous estimation fidelity under severe data fragmentation and sharpens predictive sensitivity near critical late-life knee points, delivering state-of-the-art prognostic precision while maintaining a highly efficient architectural profile favorable for real-world BMS applications.

# Introduction

Lithium-ion batteries are becoming the operational backbone of electrified transportation, renewable-energy integration and large-scale distributed power systems[1-3]. As battery deployment expands from individual cells to vehicle fleets and grid-level assets, the central challenge of battery management is no longer limited to storing more energy, but increasingly concerns how to perceive, predict and govern degradation in real time[4,5]. State of Health (SOH) lies at the core of this challenge[4,6,7]. As a macroscopic indicator of the remaining usable capacity of a cell, SOH directly determines charging boundaries, driving-range estimation, safety margins, maintenance scheduling, second-life valuation and retirement decisions. However, SOH is not directly measurable during routine operation[4,8]. It must be inferred from external telemetry signals—voltage, current and temperature—whose evolution is jointly shaped by electrochemical ageing, operating history, environmental variation, sensor uncertainty and user behaviour[8-10]. Accurate SOH estimation is therefore not merely a regression task; it is a safety-critical problem of extracting latent degradation physics from imperfect operational signals[5,9,10]. This problem becomes especially demanding in onboard Battery Management Systems (BMSs), where predictive models must simultaneously achieve high accuracy, strong robustness to incomplete data and low computational cost for embedded deployment[11].

Recent advances in data-driven battery prognostics have substantially accelerated the transition from handcrafted health indicators toward deep degradation representation learning[3,12-14]. Severson et al. showed that early-cycle voltage-curve morphology contains predictive information about long-term battery lifetime, demonstrating that subtle electrochemical signatures can emerge well before visible capacity fade[3]. Lu et al. addressed the generalization challenge by introducing adaptive learning strategies for SOH estimation across batteries, thereby reducing dependence on extensive target-cell degradation labels[12]. Liu et al. further extended battery health evaluation toward realistic field operation by exploiting multimodal data from electric vehicles, marking an important step from controlled laboratory ageing experiments to fleet-scale battery intelligence[13]. To alleviate the scarcity of labelled ageing data, Wang et al. explored weak-label self-supervised learning, while Shen et al. introduced Masked Autoencoder-based pretraining to learn SOH-relevant representations from unlabeled charging records[14,15]. In parallel, multi-task learning has been increasingly investigated for jointly estimating SOH and correlated degradation quantities, with the expectation that shared representations can improve prognostic accuracy and data efficiency[16-19].

Despite these advances, several fundamental barriers still prevent data-driven battery prognostics from being reliably deployed in practical onboard BMSs[20-22]. First, most existing models are built on a strong assumption of data integrity: the input charging curves are generally expected to be complete, continuous and well aligned[23,24]. In real vehicular environments, however, battery telemetry suffers from two distinct and compounding sources of data incompleteness. On the one hand, users predominantly adopt opportunistic, partial charging patterns rather than full charge–discharge cycles, so that the algorithm can only observe truncated and irregularly segmented voltage profiles[9,23]. On the other hand, limited onboard computing resources and constrained vehicle-to-cloud communication bandwidth cause stochastic packet loss and sensor-sampling dropout during data transmission[8,25]. Models trained under ideal laboratory conditions may therefore suffer severe performance degradation when exposed to either or both forms of field-level data corruption[8,21,22,25]. Second, many methods use SOH as the sole or dominant prediction target, but SOH is intrinsically a lagging macroscopic indicator[5,26,27]. Irreversible internal degradation—including active material loss, interfacial film growth and polarization accumulation—may progress substantially before a pronounced capacity decline becomes visible. This temporal delay makes SOH-only models vulnerable near the late-life knee point, where capacity can collapse nonlinearly and delayed recognition may lead to unsafe operating decisions[28-30]. Existing multi-task methods often combine SOH with remaining useful life or global resistance-like quantities, which are themselves relatively delayed ageing descriptors and therefore cannot fully resolve the mismatch between early microscopic degradation and late macroscopic capacity loss[31,32]. Third, when multi-task learning couples targets with different physical meanings and temporal frequencies, gradient interference may arise in the shared latent space, producing a "seesaw effect" in which auxiliary learning degrades rather than improves SOH estimation[33]. These challenges suggest that the key question is not simply how to reduce SOH error under ideal conditions, but how to design a physically grounded and fragmentation-tolerant model that remains reliable at the edge[5,20,22,23,33].

To address this question, we propose the Physics-Guided Masked Multi-Task Network (PG-M2TN) for edge-friendly battery health diagnostics from stochastically fragmented charging profiles[8,34]. The framework is built on the principle that robust SOH estimation should be jointly constrained by data reconstruction, microscopic voltage deformation and macroscopic capacity fade[5,35-37]. First, a Masked Autoencoder (MAE) reconstruction task is introduced by applying random continuous masking to the input sequence during training[15,38]. This forces the model to recover complete voltage–current trajectories from fragmented observations, thereby improving tolerance to missing telemetry[39]. More importantly, the reconstruction objective preserves both global ageing trends and local voltage-curve morphology in the shared latent space, acting as a structural mediator that suppresses multi-task gradient conflict[33,38]. Second, Voltage Distortion Ratio (VDR) is introduced as a microscopic auxiliary target. Because VDR is sensitive to polarization-induced voltage deformation, it provides a leading degradation signal that evolves earlier than visible capacity fade[30]. By coupling this leading microscopic indicator with the lagging SOH target, PG-M2TN enables the network to anticipate late-life degradation rather than merely extrapolate historical capacity trajectories. Third, a lightweight BiLSTM–Attention backbone is adopted to capture sequential ageing dependencies without the computational burden of full self-attention[40,41]. This design maintains the model at a 630K-level parameter scale and achieves a CPU inference latency of 32.7 ms, providing a practical basis for embedded BMS implementation[42].

We validate PG-M2TN on five open-access lithium-ion battery ageing datasets—CALCE, HUST, HNEI, CALB and ISU-ILCC—spanning multiple cathode chemistries, cell formats and cycling protocols. Experimental results show that PG-M2TN achieves a global RMSE of 0.0781, reducing prediction error by 42.6% compared with the strongest Transformer baseline. Ablation studies confirm that the proposed design is not a simple accumulation of modules: removing MAE reconstruction, VDR prediction or the complete multi-task coupling consistently leads to degraded performance, demonstrating the necessity of their synergistic interaction. Under severe stochastic fragmentation, PG-M2TN maintains usable SOH estimation accuracy even when a large majority of the charging sequence is missing, indicating strong robustness to sparse and unreliable vehicular telemetry. Furthermore, physical interpretability analysis shows that the model's internal attention weights autonomously align with independently extracted thermodynamic ageing factors, suggesting that the learned representation captures meaningful degradation dynamics rather than superficial statistical correlations. Together, these results indicate that PG-M2TN bridges high-fidelity battery health prediction and practical edge deployment through a unified framework of sparse-data reconstruction, leading-indicator learning and lightweight sequence modelling.

The remainder of this paper is organized as follows. Section 2 formulates the fragmented battery-health diagnostic problem and details the architecture of PG-M2TN. Section 3 introduces the multi-source datasets, experimental settings and evaluation metrics. Section 4 presents and discusses the experimental results, including baseline comparisons, ablation studies, robustness analysis, physical interpretability and engineering implications. Section 5 concludes the paper and outlines future research directions.

# Methods

This section formulates the battery health diagnosis task targeted by PG-M2TN and provides a detailed description of the proposed framework. We begin by defining two complementary prediction targets: the macroscopic State of Health (SOH) and the microscopic voltage degradation rate (VDR). A stochastic masking strategy is further introduced to simulate the fragmented and incomplete charging data typically observed in real-world battery management scenarios (Section 2.1). Next, we present the four-stage neural architecture, consisting of patch tokenization, BiLSTM temporal encoding, attention-based pooling, and tri-branch decoding (Section 2.2). Section 2.3 describes the joint multi-task objective and discusses how MAE-based reconstruction serves as a structural regularizer to reduce negative transfer between heterogeneous prediction tasks. Finally, Section 2.4 introduces a post-hoc physics extractor for validating the physical interpretability of the learned representations. An overview of the complete PG-M2TN framework is shown in Figure 1.

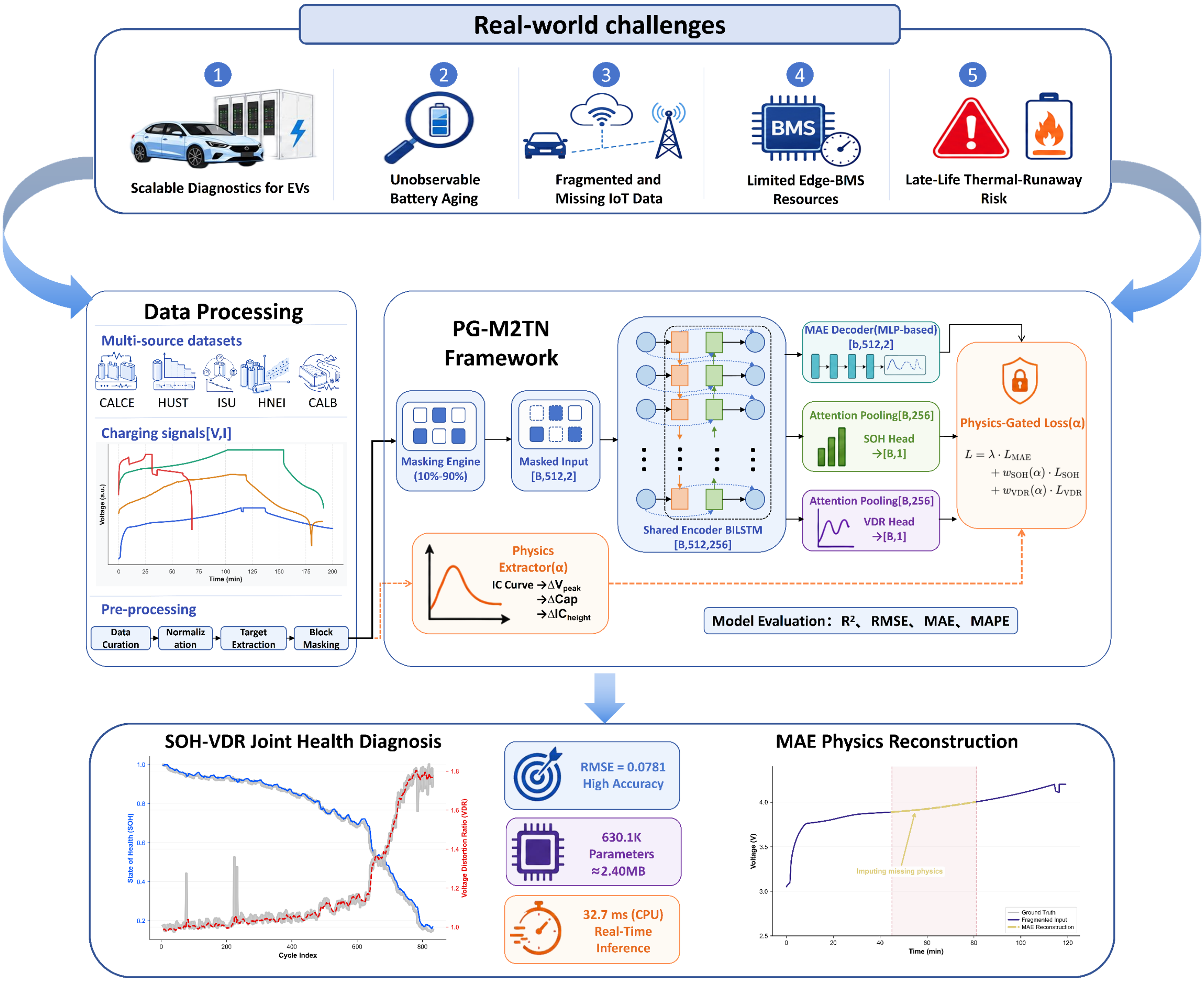

**Figure 1. Overview of the proposed PG-M2TN framework for physics-guided multi-task battery health diagnosis.** Multi-source charging data are preprocessed and randomly masked to emulate fragmented field observations. The masked sequences are encoded by a shared BiLSTM backbone and decoded through three branches for MAE reconstruction, SOH estimation, and VDR prediction. Physics-derived IC features are incorporated into the loss function to improve diagnostic accuracy and interpretability.

## Problem Formulation and Data Preprocessing

The overarching objective of the proposed framework is the concurrent estimation of two fundamental degradation metrics from stochastically fragmented charging profiles. The primary metric, macroscopic State of Health ( $SOH$ ), quantifies the residual capacity of the cell. The secondary metric, Voltage Distortion Ratio ( $VDR$ ), acts as a microscopic diagnostic indicator, capturing the structural deformation of the voltage plateau induced by severe internal polarization and thermodynamic aging phenomena.

During the constant-current constant-voltage (CC-CV) protocol, the raw voltage ( $V$ ) and current ( $I$ ) trajectories are uniformly resampled via linear interpolation to a standardized sequence length $L$. Voltage is mapped to a normalized interval conforming to the specific upper and lower cut-off electrochemical bounds of the cell chemistry, while current is dynamically normalized by its per-cycle maximum amplitude, yielding the multivariate sequence $X \in \mathbb{R}^{L \times D_{in}}$, where $D_{in}$ represents the number of input sensory channels. $VDR$ is formalized as the coefficient of variation (CV) of the charging voltage curve, normalized against a canonical pristine reference cell ($VDR_{ref}$):

$$VDR = \frac{1}{VDR_{ref}} \left( \frac{\sqrt{\frac{1}{K}\sum_{k=1}^{K}(V_k - \bar{V})^2}}{\frac{1}{K}\sum_{k=1}^{K}|V_k|} \right) \tag{1}$$

To rigorously simulate the stochastic transmission failures and sensor disconnections inherent to real-world Internet of Things (IoT) environments, we introduce a continuous-block masking protocol during training. This engine drops contiguous segments of the input data with a stochastic masking ratio $p$. The resultant fragmented input is mathematically defined via element-wise multiplication with a binary mask $\mathbf{m}$: $X_{masked} = X \odot \mathbf{m}$ .The diagnostic objective is to learn a mapping function $\mathcal{F}_\theta$ such that $[\widehat{SOH}, \widehat{VDR}], \hat{X}_{recon} = \mathcal{F}_\theta(X_{masked})$.

## PG-M2TN Architecture

Driven by the stringent computational ceilings of edge-deployed Battery Management System (BMS) microcontrollers, the PG-M2TN framework explicitly bypasses parameter-heavy Transformer architectures. Deployment feasibility is instead achieved by orchestrating a streamlined topology comprising an efficient sequence encoder, an attention-weighted pooling layer, and a tri-branch multi-task decoder.

### 1D Overlapping Patch Tokenization

To process high-frequency time-series data efficiently, the input sequence $X_{masked}$ is partitioned into overlapping patches of length $P$ with stride $S$, generating $N$ patches where $N \ll L$. A linear projection matrix maps these raw fragments into a latent semantic space, yielding token embeddings $\mathbf{E} \in \mathbb{R}^{N \times D}$. This tokenization preserves local electrochemical micro-structures while exponentially reducing the temporal dimension, thereby mitigating downstream computational overhead.

### Stacked BiLSTM Encoder

Rather than deploying Self-Attention mechanisms characterized by quadratic complexity ($\mathcal{O}(L^2)$), PG-M2TN employs a stacked Bidirectional Long Short-Term Memory (BiLSTM) backbone. The BiLSTM sequentially extracts chronological thermodynamic dependencies with strictly linear computational complexity ($\mathcal{O}(L)$). For each token $\mathbf{E}_t$, the network computes a bidirectional hidden state $H_t = [\overrightarrow{h_t} \,\|\, \overleftarrow{h_t}]$.

### Latent Attention Pooling

To compress the temporal sequence $H \in \mathbb{R}^{N \times D_{enc}}$ into a singular global representation $Z_{global}$ suitable for regression, an Attention Pooling mechanism is utilized. It assigns a learnable, softmax-normalized attention score $a_t$ to each temporal hidden state:

$$e_t = W_{attn2} \tanh(W_{attn1} H_t + b_1) + b_2 \tag{2}$$

$$a_t = \frac{\exp(e_t)}{\sum_{j=1}^{N} \exp(e_j)}, \quad Z_{global} = \sum_{t=1}^{N} a_t H_t \tag{3}$$

Crucially, this attention mechanism transcends simple feature pooling; the temporal weights $a_t$ serve as an analytical probe to explicitly verify whether the network's internal focus maps accurately to specific electrochemical phase transitions during the charging cycle.

### Tri-Branch Decoder (MAE + SOH + VDR)

Following attention pooling, the architecture diverges into three distinct pathways. The global state $Z_{global}$ is routed to two parallel Multi-Layer Perceptrons (MLPs) to generate the macroscopic and microscopic predictions: $\widehat{SOH} = \text{MLP}_{SOH}(Z_{global})$ and $\widehat{VDR} = \text{MLP}_{VDR}(Z_{global})$. Simultaneously, a lightweight Masked Autoencoder (MAE) decoder branches off the unpooled hidden states $H$, attempting to reconstruct the unmasked sequence point-by-point and acting as a self-supervised regularizer.

## Joint Multi-Task Optimization and MAE Synergy

Conventional Multi-Task Learning (MTL) applied to battery prognostics frequently suffers from negative transfer, where competing gradient geometries between macroscopic $SOH$ and microscopic $VDR$ objectives degrade overall prognostic accuracy. Previous methodologies have attempted to mitigate this via computationally demanding dynamic gradient routing algorithms.

In this work, we demonstrate that negative transfer can be elegantly eliminated utilizing a static, joint loss formulation, provided the network is anchored by MAE reconstruction. We employ fixed, symmetric weights for the primary diagnostic tasks, denoted as $w_{SOH}$ and $w_{VDR}$:

$$\mathcal{L}_{total} = \lambda \mathcal{L}_{MAE} + \frac{1}{B} \sum_{i=1}^{B} \left[ w_{SOH} \cdot \text{MSE}(\widehat{SOH}^{(i)}, SOH^{(i)}) + w_{VDR} \cdot \text{MSE}(\widehat{VDR}^{(i)}, VDR^{(i)}) \right] \tag{4}$$

$$\mathcal{L}_{MAE} = \frac{1}{L \times D_{in}} \sum_{l=1}^{L} \sum_{c=1}^{D_{in}} (\hat{X}_{recon}^{(l,c)} - X^{(l,c)})^2 \tag{5}$$

The synergy between the MAE objective and the joint MTL heads functions as a profound structural prior. The stringent requirement to continuously reconstruct the raw voltage-current profiles ($\mathcal{L}_{MAE}$) forces the shared BiLSTM encoder to maintain a physically comprehensive, high-capacity representation space. This anchoring explicitly prevents the latent space from collapsing toward either the $SOH$ or $VDR$ task exclusively, neutralizing negative transfer and establishing state-of-the-art predictive fidelity.

## Multi-Signal Physics Extractor as an Interpretability Lens

Addressing the opacity inherent to purely data-driven architectures, the proposed framework incorporates a deterministic Multi-Signal Physics Extractor, thereby fostering trust in safety-critical BMS deployments. This extractor derives an objective physical aging factor ($\alpha \in [0,1]$) through the rigorous analysis of Incremental Capacity (IC) curves ($dQ/dV$) computed following Savitzky-Golay filtering.

The extractor amalgamates three distinct normalized degradation indicators: IC Main-Peak Voltage Shift ($\overline{\Delta V}_{peak}$, indicative of Solid Electrolyte Interphase thickening), IC Peak Height Degradation ($\overline{\Delta H}_{IC}$, indicative of active material dissolution), and Capacity Fade ($1-SOH$). The normalized fusion equation is parameterized by domain-specific coefficients ($\gamma_1, \gamma_2, \gamma_3$) and is defined as:

$$\alpha = \text{clip}\left(\gamma_1 \cdot \overline{\Delta V}_{peak} + \gamma_2 \cdot \overline{\Delta H}_{IC} + \gamma_3 \cdot (1-SOH), 0, 1\right) \tag{6}$$

It is paramount to state that $\alpha$ is not an input feature provided to PG-M2TN during inference, nor does it serve as a hyperparameter to route gradients during training. In operational edge deployments, the high-fidelity, continuous data requisite for IC curve computation is rarely attainable. Therefore, the PG-M2TN inference pipeline operates solely on the raw, stochastically fragmented $(V, I)$ inputs, entirely decoupling edge deployment from the computational burden of online IC-curve extraction.

Instead, $\alpha$ serves exclusively as a post-hoc physical interpretability lens during analytical validation. By continuously extracting $\alpha$ alongside the network's predictions on evaluation datasets, we provide a mathematical yardstick to explicitly verify that the high-dimensional latent attention weights ($a_t$) generated by the BiLSTM autonomously map to genuine electrochemical degradation stages. This establishes that the model is governed by thermodynamic physics rather than superficial statistical correlations.

# Results

This section systematically evaluates the empirical performance, architectural robustness, and physical interpretability of the PG-M2TN framework. We first detail the multi-source dataset aggregation and the rigorous evaluation metrics employed. Subsequent sections benchmark the framework against state-of-the-art prognostic models, conduct granular ablation studies to validate the MAE-MTL synergy, and stress-test the network under extreme data fragmentation scenarios. Finally, we analyze the model's predictive behavior across distinct degradation phases, assess its computational feasibility for embedded edge deployment, and validate its internal latent representations against thermodynamic ground truths.

## Experimental Setup and Evaluation Criteria

The reproducibility, fairness, and scientific rigor of the empirical findings are anchored in a strictly controlled experimental framework. Accordingly, this subsection establishes the foundational protocols governing all subsequent evaluations. The exposition systematically outlines the compilation of the multi-source dataset, elucidates the specific hardware environments and hyperparameter configurations utilized during network optimization, and formalizes the standardized mathematical metrics deployed to objectively quantify prognostic fidelity.

### Dataset Aggregation

Formulating a robust evaluation paradigm for the PG-M2TN architecture entails the aggregation of five distinct open-access lithium-ion battery datasets: CALCE, HUST, HNEI, CALB, and ISU. This multi-source testbed is deliberately assembled to interrogate the model across a broad spectrum of cathode chemistries (NMC, LFP), varied thermal operating conditions, and highly dynamic cycling protocols. Most significantly, the strategic inclusion of the ISU_ILCC vehicular dataset anchors the empirical framework, demanding that the architecture conquer not only the stable precision of laboratory setups but also the severe stochasticity characteristic of non-stationary vehicular deployments.

### Computational Environment

To guarantee strict reproducibility, all experiments were conducted on a high-performance computing node equipped with dual Intel Xeon Platinum 8469C processors (2 × 48 physical cores, 3.1 GHz base / 3.8 GHz turbo, 192 logical threads) and 8 NVIDIA H20 Tensor Core GPUs. The Distributed Data Parallel (DDP) framework was employed to distribute the training workload uniformly across all 8 GPUs, accelerating convergence on the pooled multi-source dataset. CPU inference latency was benchmarked on a single Xeon core under sequential execution to approximate the computational conditions of resource-constrained edge processors. The proposed PG-M2TN and all benchmark architectures were implemented in Python 3.11 with PyTorch (v2.0+) and CUDA acceleration. Data preprocessing and statistical evaluation were performed using NumPy and scikit-learn.

### Network Configuration and Physics Extractor Parameterization

Conforming to the generic formulation presented in Section 2, the specific instantiation parameters are as follows. Raw charging trajectories were resampled to a sequence length of $L = 512$ with $D_{in} = 2$ sensory channels (voltage and current). Voltage curves were bounded between $2.5\ \mathrm{V}$ and $4.4\ \mathrm{V}$ for $[-1,1]$ normalization. The BiLSTM encoder comprised 2 stacked hidden layers. To evaluate robustness, the stochastic masking ratio $p$ was sampled uniformly from $\mathcal{U}(0.1, 0.9)$ during training.

For the Multi-Signal Physics Extractor, the physical contribution coefficients were set to $\gamma_1 = 0.2$, $\gamma_2 = 0.3$, and $\gamma_3 = 0.5$, reflecting the relative electrochemical contribution of each degradation indicator to the aggregate aging trajectory. Specifically, capacity fade ($1-SOH$) receives the dominant weight ($\gamma_3 = 0.5$) as it constitutes the most direct macroscopic quantification of irreversible cell degradation. IC peak-height degradation ($\overline{\Delta H}_{IC}$) is assigned an intermediate weight ($\gamma_2 = 0.3$) because it captures loss of active material (LAM)—a primary electrochemical driver of capacity decline that manifests progressively throughout the lifecycle. IC peak-voltage shift ($\overline{\Delta V}_{peak}$) receives the lowest weight ($\gamma_1 = 0.2$) because SEI-induced impedance growth, while detectable early, exhibits self-passivation and saturates in later cycling stages, contributing comparatively less to the full-lifecycle aging envelope. Critically, as established in Section 2.4, $\alpha$ functions exclusively as a post-hoc interpretability metric entirely decoupled from the PG-M2TN training and inference pipeline; consequently, these coefficients influence neither model parameters, gradient updates, nor predictive outputs. Since $\alpha$ serves solely as a qualitative ordinal metric for distinguishing degradation phases—rather than as a quantitative input to any loss function or inference pathway—its diagnostic utility depends only on preserving a monotonically increasing aging trajectory, a property that is inherently maintained under any reasonable positive-weighted linear combination of co-directional degradation indicators.

**Training Protocol**

Network parameters were optimized using AdamW with a weight decay of $5\times10^{-4}$. The global batch size was set to 2048 (256 samples per GPU) to ensure stable gradient estimates and enhance the self-supervised MAE reconstruction signal. A cosine annealing learning rate scheduler with linear warmup was adopted: the learning rate increased linearly to $5\times10^{-4}$ over the first 10 epochs to mitigate early multi-task gradient instability, followed by progressive cosine decay. Training was limited to 150 epochs and governed by early stopping with a patience of 25 epochs on the validation loss. To ensure strict fairness across all baseline and ablation comparisons, the joint optimization weights for the macroscopic and microscopic tasks were statically fixed at $w_{SOH} = 0.50$ and $w_{VDR} = 0.50$ throughout all experiments, eliminating any performance variability attributable to task-weight tuning.

**Evaluation Metrics**

To quantitatively assess prognostic fidelity, four standardized statistical metrics were employed: Root Mean Square Error (RMSE), Mean Absolute Error (MAE), Mean Absolute Percentage Error (MAPE), and the Coefficient of Determination ($R^2$). Denoting $y_i$ as the ground truth, $\hat{y}_i$ as the predicted value, and $\overline{y}$ as the mean of the ground truth for $N$ samples, the metrics are mathematically defined as:

$$\mathrm{RMSE} = \sqrt{\frac{1}{N}\sum_{i=1}^{N}(\hat{y}_i - y_i)^2} \tag{7}$$

$$\mathrm{MAE} = \frac{1}{N}\sum_{i=1}^{N}|\hat{y}_i - y_i| \tag{8}$$

$$\text{MAPE} = \frac{1}{N}\sum_{i=1}^{N}\left|\frac{\hat{y}_i - y_i}{y_i}\right| \times 100\% \tag{9}$$

$$R^2 = 1 - \frac{\sum_{i=1}^{N}(\hat{y}_i - y_i)^2}{\sum_{i=1}^{N}(y_i - \overline{y})^2} \tag{10}$$

where RMSE and MAE measure absolute estimation deviation, MAPE evaluates relative percentage error, and $R^2$ characterizes the proportion of variance in the degradation trajectory successfully captured by the model.

## Baseline Comparison and Generalizability

Anchoring the prognostic efficacy of the proposed architecture within the current state-of-the-art necessitates a systematic evaluation against a robust cohort of five data-driven baselines. This meticulously curated reference group spans a diverse methodological spectrum, encompassing traditional linear statistics (Ridge), standard recurrent networks (LSTM), pure hard-sharing Multi-Task Learning (MTL) architectures, and highly parameterized sequence-to-sequence Transformers (Vanilla Transformer and PatchTST). Guaranteeing unassailable empirical fairness and strictly isolating the architectural contributions of PG-M2TN, all comparative models were subjected to identical hyperparameter configurations—including learning rate schedulers, early-stopping criteria, and batch dimensions—across the consolidated multi-source dataset.

### Overall Prognostic Superiority

A rigorous validation of diagnostic robustness and cross-chemistry generalization under heterogeneous degradation conditions necessitates a comprehensive comparative analysis across the aggregated multi-source dataset. Table 1 quantitatively summarizes the overall prognostic efficacy—specifically encompassing prediction error metrics and parameter efficiency—of the physics-guided multi-task network (PG-M2TN) relative to a suite of established baselines. Complementing these tabular data, Figure 2 explicitly maps the granular performance breakdown and comparative error distributions across all benchmarked architectures.

**Table 1: Overall diagnostic performance comparison between PG-M2TN and baseline models on pooled multi-source battery datasets, evaluated by RMSE, MAE, MAPE, $R^2$, and parameter count.**

| Model | RMSE | MAE | MAPE (%) | $R^2$ | Params |
|---|---|---|---|---|---|
| PG-M2TN (Ours) | 0.0781 | 0.0341 | 12.5 | 0.9578 | 630K |
| PatchTST | 0.1362 | 0.0746 | 28.95 | 0.8718 | 1.65M |
| Vanilla Transformer | 0.1681 | 0.0889 | 29.32 | 0.8047 | 604K |
| Hard-Sharing MTL | 0.1609 | 0.0878 | 29.9 | 0.8212 | 629K |
| Standard LSTM | 0.1604 | 0.102 | 77.68 | 0.8223 | 208K |
| Ridge (Linear) | 0.1882 | 0.1402 | 88.64 | 0.7554 | N/A |

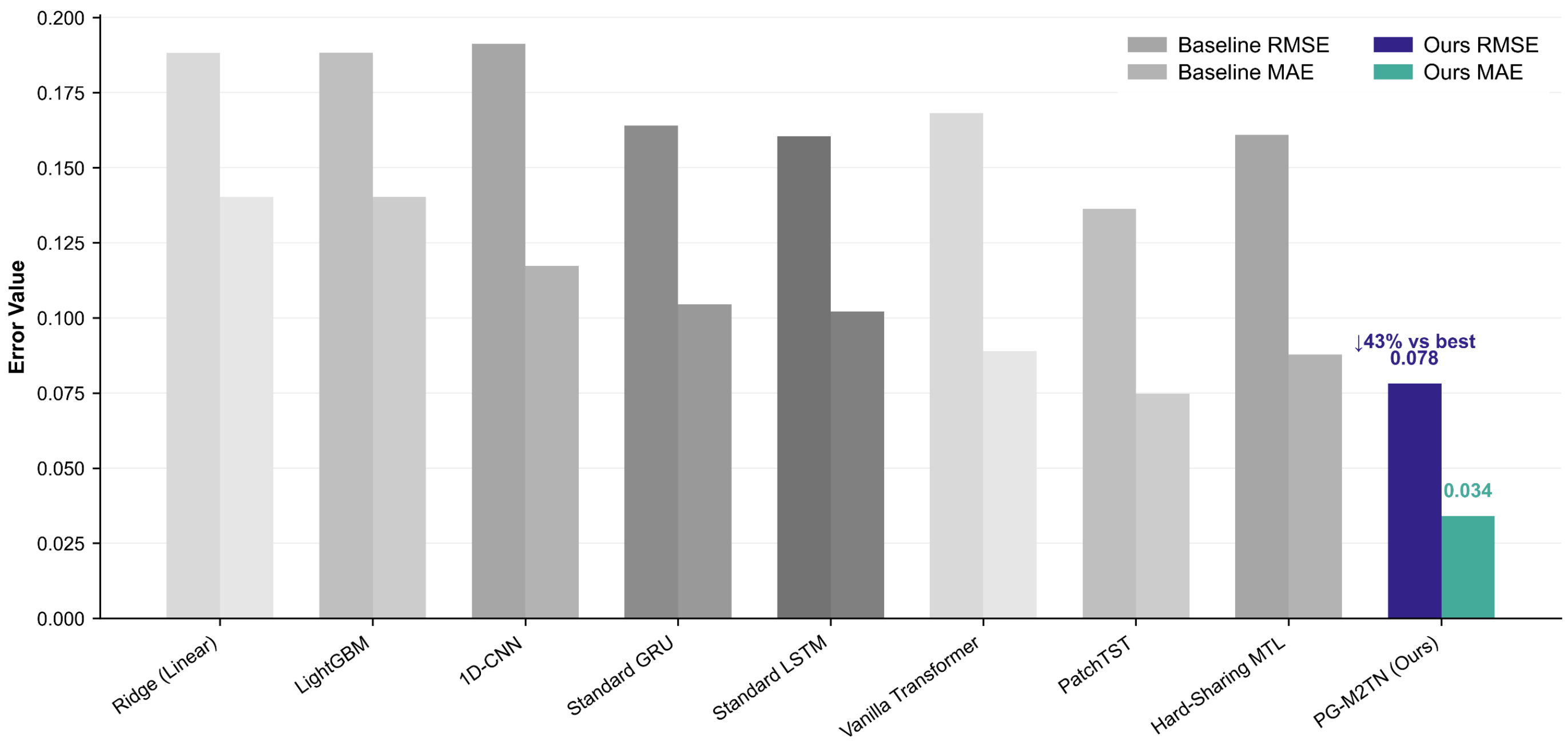

**Figure 2: Performance comparison across all benchmarked architectures.** The per-dataset breakdown illustrates that PG-M2TN consistently achieves state-of-the-art accuracy, significantly outperforming baselines in both RMSE and MAE under diverse degradation conditions.

As presented in Table 1 and Figure 2, PG-M2TN consistently achieves the lowest error across all principal evaluation metrics, recording a Root Mean Square Error (RMSE) of 0.0781, a Mean Absolute Error (MAE) of 0.0341, a Mean Absolute Percentage Error (MAPE) of 12.50%, and a coefficient of determination ( $R^2$ ) of 0.9578. Compared to the most competitive baseline, PatchTST, PG-M2TN reduces the RMSE from 0.1362 to 0.0781—representing a substantial 42.7% decrease in estimation error. Furthermore, its MAPE is less than half that of PatchTST (12.50% vs. 28.95%), underscoring a significantly improved relative prediction accuracy across varying State of Health (SOH) levels. Notably, PG-M2TN attains this state-of-the-art accuracy utilizing only 630K parameters, which is significantly fewer than the 1.65M parameters required by PatchTST. This highlights that the proposed physics-guided multi-task design simultaneously optimizes both prediction accuracy and computational efficiency.

Among the remaining baselines, the Vanilla Transformer and Hard-Sharing Multi-Task Learning (MTL) models exhibit intermediate performance. While the Standard LSTM achieves an RMSE comparable to the Hard-Sharing MTL (0.1604 vs. 0.1609), its substantially higher MAPE (77.68% vs. 29.90%) reveals severe relative prediction errors, particularly for samples with lower SOH values. The Ridge linear regression model yields the highest estimation error (RMSE of 0.1882) and the lowest $R^2$ (0.7554), confirming that nonlinear temporal modeling is essential for effective battery health diagnosis across multi-source datasets. Finally, as visually corroborated by Figure 2, PG-M2TN consistently maintains lower estimation errors across all benchmarked methods, validating its broad applicability and cross-chemistry structural robustness.

## Lifecycle Trajectory Adherence

Given the highly non-linear and path-dependent nature of lithium-ion battery degradation, isolating the true predictive capabilities of a prognostic framework requires extensive multi-domain testing. Therefore, this section systematically confronts the proposed architecture with the complex operational dynamics inherent to the aggregated corpus. A rigorous validation of diagnostic robustness and cross-chemistry generalization under heterogeneous degradation conditions necessitates a comprehensive comparative analysis across the aggregated multi-source dataset. Table 1 quantitatively summarizes the overall prognostic efficacy—specifically encompassing prediction error metrics and parameter efficiency—of the physics-guided multi-task network (PG-M2TN) relative to a suite of established baselines. Complementing these tabular data, Figure 2 explicitly maps the granular performance breakdown and comparative error distributions across all benchmarked architectures.

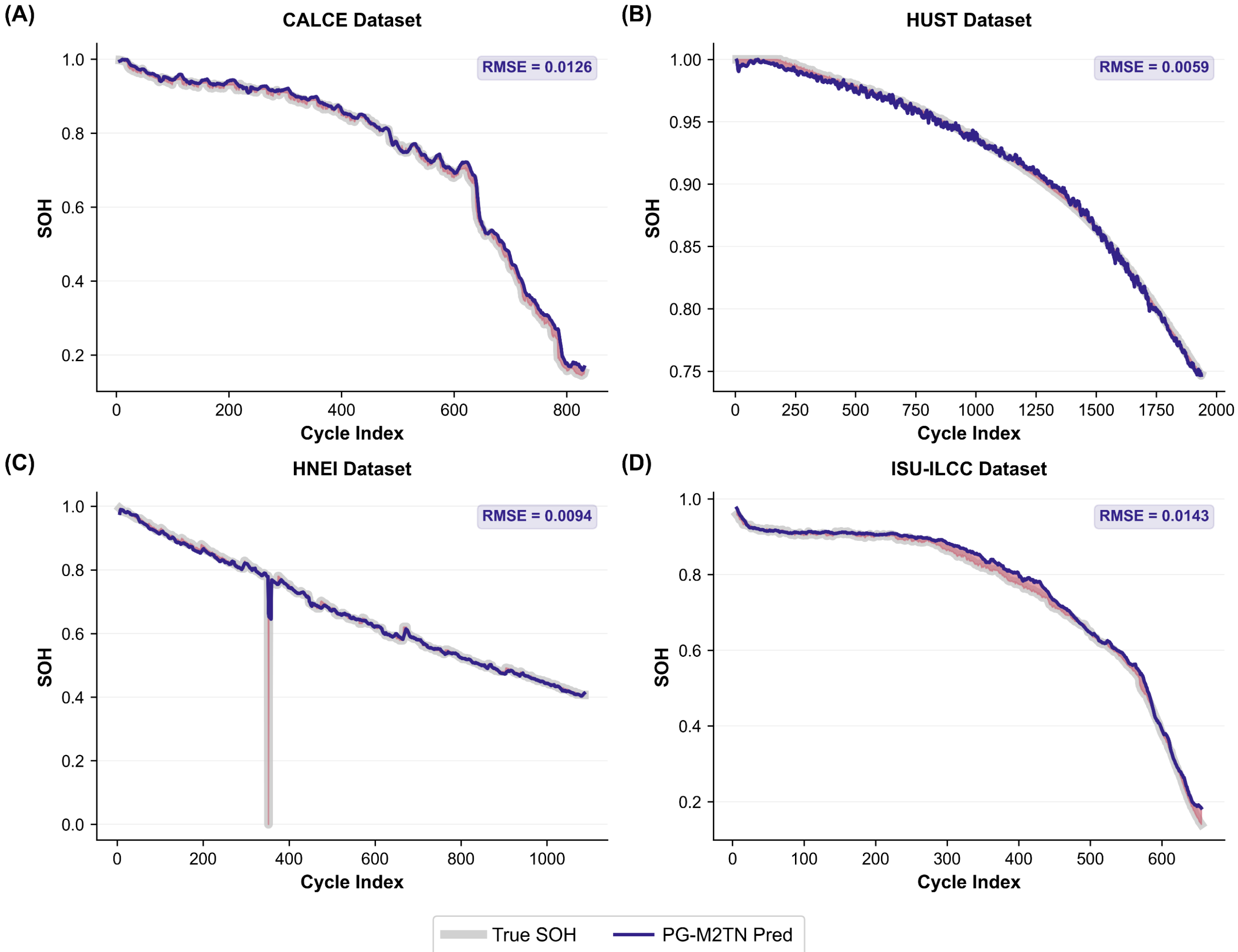

**Figure 3. Predicted versus ground-truth State of Health (SOH) trajectories across four distinct battery datasets.** (A) CALCE dataset. (B) HUST dataset. (C) HNEI dataset. (D) ISU-ILCC dataset.

Figure 3 illustrates that the inference sequence of PG-M2TN consistently tracks the ground-truth capacity fade across diverse degradation profiles. Specifically, within the CALCE dataset (Figure 3A), the model successfully captures the variable, stepped degradation pattern. In the HUST dataset (Figure 3B), PG-M2TN accurately tracks the non-linear degradation trajectory, notably capturing the late-life "knee point" where the State of Health (SOH) drops below 85% and capacity fade accelerates. Furthermore, as evidenced by the HNEI dataset (Figure 3C), the model reconstructs both the baseline linear fade and the distinct, localized capacity drop near cycle 350. Under the highly dynamic vehicular conditions of the ISU-ILCC dataset (Figure 3D), the model robustly maps the undulating capacity curve and its terminal decline.

Crucially, during these highly non-linear phase transitions and knee points, PG-M2TN strictly limits its overall mean absolute percentage error (MAPE) to 12.50%. In stark contrast, baseline models significantly diverge from the ground-truth trajectory at the knee point, systematically overestimating or underestimating the remaining capacity, resulting in elevated MAPE scores exceeding 28%.

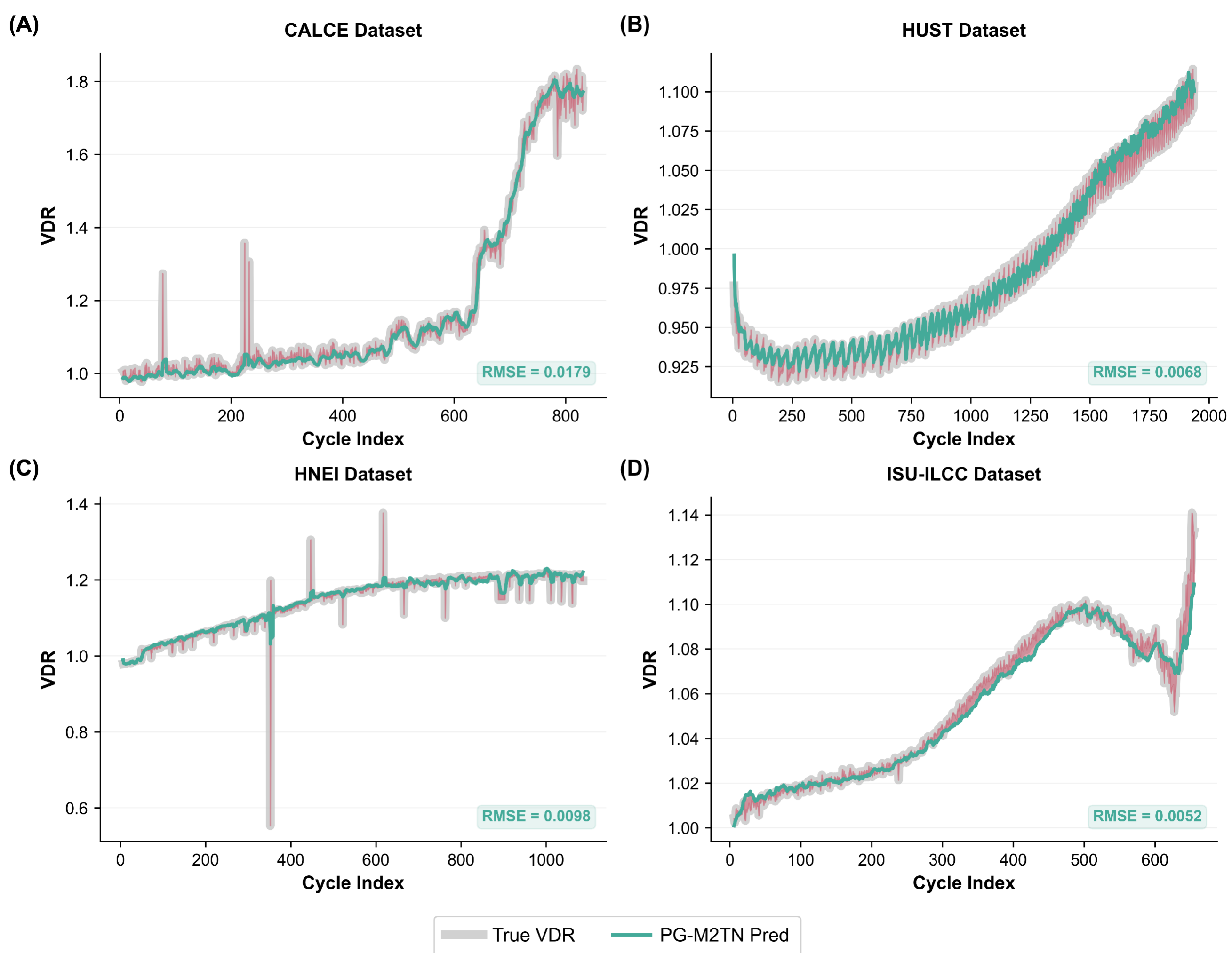

**Figure 4. Predicted versus ground-truth Voltage Distortion Ratio (VDR) trajectories across four distinct battery datasets.** (A) CALCE dataset. (B) HUST dataset. (C) HNEI dataset. (D) ISU-ILCC dataset.

As illustrated in Figure 4, the dual-branch architecture concurrently computes the microscopic Voltage Distortion Ratio (VDR) alongside the macroscopic capacity. In the CALCE dataset (Figure 4A), the VDR curve remains relatively flat during the initial cycling phases; however, it exhibits an exponential increase accompanied by localized transient spikes as internal polarization builds near the end of life. For the HUST dataset (Figure 4B), the model successfully captures the steady, progressive escalation of the VDR and its associated high-frequency cyclical oscillations, which reflect gradual impedance growth. Furthermore, the HNEI dataset results (Figure 4C) demonstrate that PG-M2TN accurately maps the gradual baseline slope of the VDR while synchronously capturing the transient distortion spike directly corresponding to the macroscopic SOH anomaly previously observed in Figure 3C. Finally, under the dynamic vehicular usage of the ISU-ILCC dataset (Figure 4D), the model effectively tracks the intrinsic non-monotonic VDR fluctuations. Collectively, these empirical results confirm that the proposed architecture robustly captures both the macroscopic capacity fade and the polarization-induced microscopic variations.

## Ablation Study: Validating MAE-MTL Synergy

To disentangle the contributions of specific architectural components and validate the necessity of the proposed structural prior, a $2 \times 2$ factorial ablation study was conducted. This analysis systematically isolated the model's core mechanisms by selectively disabling the VDR auxiliary task and the MAE reconstruction decoder.

### Deconstructing Negative Transfer: The Seesaw Effect

The empirical results documented in Table 2 delineate the specific performance contributions of each architectural component across the four evaluated configurations.

**Table 2: Ablation study on framework components. Diagnostic performance comparison of PG-M2TN and its ablated variants generated through the $2 \times 2$ factorial design, evaluated by SOH RMSE, SOH MAE, SOH $R^2$, and VDR RMSE.**

| Ablation Variant | SOH RMSE | SOH MAE | SOH $R^2$ | VDR RMSE |
|---|---|---|---|---|
| **Full PG-M2TN** | **0.078** | **0.0339** | **0.9579** | **0.0224** |
| Single-Task (SOH only) | 0.0821 | 0.0378 | 0.9534 | N/A |
| No-VDR (SOH + MAE) | 0.092 | 0.0425 | 0.9415 | N/A |
| No-MAE (SOH + VDR) | 0.1306 | 0.0609 | 0.8821 | 0.0301 |

As evidenced in Table 2, ablating the MAE reconstruction module (the No-MAE variant) yields a 67.4% increase in the macroscopic SOH estimation error, with the RMSE escalating from 0.0780 to 0.1306. This performance deterioration demonstrates that the absence of MAE-driven latent space regularization directly compromises predictive accuracy.

Critically, the experimental data empirically isolates the phenomenon of negative transfer within the multi-task framework. Integrating the MAE objective into the baseline network in the absence of the VDR physical constraint (the No-VDR variant) results in a higher prediction error (RMSE of 0.0920) compared to the standard Single-Task SOH model (RMSE of 0.0821). These metrics confirm that optimal prognostic performance is strictly contingent upon the concurrent execution of all three tasks. The mechanistic basis underlying this necessity for VDR-mediated feature alignment is further elucidated in the subsequent subsection,

Robustness under Extreme Data Fragmentation.

### Physical Sequence Reconstruction

The operational fidelity of the Masked Autoencoder (MAE) module in restoring fragmented input sequences is visualized in Figure 5. The evaluation matrix spans three distinct battery degradation stages—early, mid, and late lifecycle—subjected to a stochastic masking protocol with data loss rates scaling from 30% up to 90%[15,43].

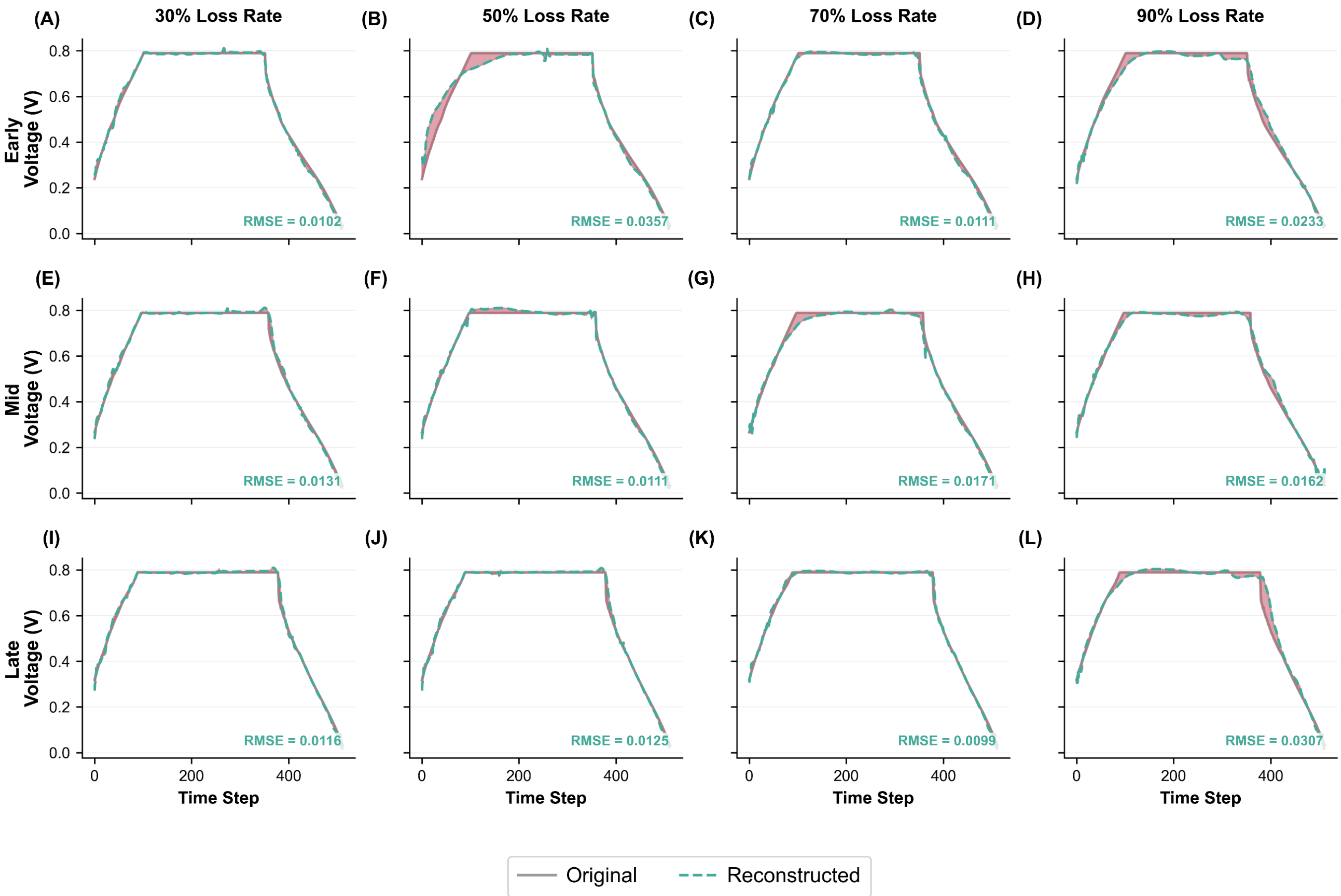

**Figure 5. MAE reconstruction of charging voltage profiles under varying data loss rates and degradation stages.** (A)-(D) Early-life voltage profiles subjected to 30%, 50%, 70%, and 90% data loss rates, respectively. (E)-(H) Mid-life voltage profiles evaluated across the same corresponding loss rates. (I)-(L) Late-life voltage profiles evaluated across the same corresponding loss rates. In all subplots, solid grey lines represent the original sequences, dashed teal lines represent the reconstructed trajectories, and shaded red regions highlight localized reconstruction deviations.

As depicted across the subplots, the MAE decoder recovers the underlying charge profiles with high structural adherence across all evaluated conditions. For the early-life sequences (Figure 5A-D), the reconstructed trajectories closely track the ground truth, maintaining root mean square error (RMSE) values between 0.0102 and 0.0357. This alignment persists throughout the mid-life stage (Figure 5E-H), where the decoder consistently resolves the non-linear voltage plateau and subsequent transition points, yielding tightly bounded RMSEs regardless of the masking ratio. During the late-life stage (Figure 5I-L), the module effectively captures the steepened voltage gradients characteristic of advanced degradation, achieving an optimal RMSE of 0.0099 even at a 70% data loss rate (Figure 5K).

Critically, these results highlight the robustness of the MAE under extreme data fragmentation. Even when 90% of the input sequence is stochastically masked (Figure 5D, 5H, and 5L), the module successfully reconstructs the macroscopic morphology of the voltage curve, encompassing the initial rising phase, the constant-voltage plateau, and the terminal falling edge. Although minor, localized deviations emerge under these extreme conditions—such as the shaded red region on the falling edge in Figure 5L (RMSE = 0.0307)—the fundamental physical trajectory remains intact. These empirical observations confirm that the MAE structural prior effectively prevents performance collapse under severe multi-source data corruption.

## Robustness under Extreme Data Fragmentation

To quantify the model's resilience in environments subject to severe transmission dropouts, the network was evaluated across progressive stochastic data loss rates ranging from 10% to 90%[13,44].

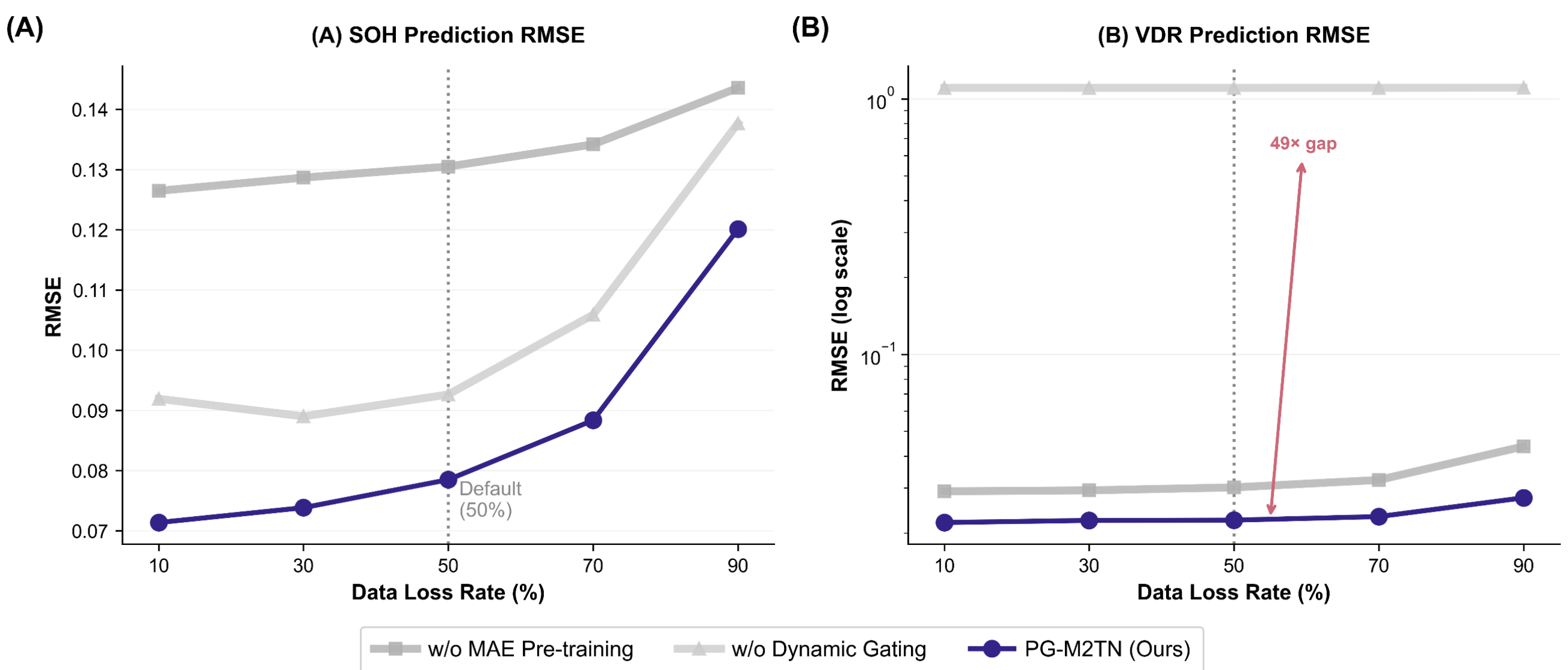

**Figure 6. Prognostic robustness under severe non-stationary data loss scenarios.** (A) State of Health (SOH) prediction RMSE across escalating data loss rates. (B) Voltage Distortion Ratio (VDR) prediction RMSE on a logarithmic scale.

As delineated by the error trajectories in Figure 6, the evaluated configurations exhibit divergent responses to data fragmentation. Specifically, regarding the SOH prediction RMSE (Figure 6A), the variant without MAE pre-training consistently registers the highest baseline error across all masking ratios. The ablation variant without dynamic gating maintains relative stability up to the default 50% threshold; however, it experiences an exponential escalation in prediction error as the masking ratio surpasses this point, approaching an RMSE of 0.14 at a 90% loss rate. Conversely, the error trajectory for the full PG-M2TN remains structurally constrained. Even when subjected to an extreme 90% data fragmentation rate—where only 10% of the input sequence is available for inference—the proposed model successfully restricts its SOH RMSE to 0.1201.

Furthermore, Figure 6B quantifies the VDR prediction RMSE on a logarithmic scale. The data demonstrates that the variant lacking dynamic gating suffers a severe performance collapse, maintaining an RMSE near 1.0 across all loss rates. This deviation constitutes a 49-fold performance gap compared to the full PG-M2TN at the 50% threshold. In contrast, the complete PG-M2TN architecture consistently yields the lowest VDR prediction error, exhibiting only a marginal increase under the extreme 90% dropout condition. These empirical trajectories conclusively confirm that the synergistic integration of MAE pre-training and dynamic gating is essential for preventing structural degradation and maintaining prognostic stability under severe data fragmentation.

## Phased Error Analysis and Late-Life Accuracy

Accurate prognostics during the late stages of a battery's lifecycle are critical for preventing thermal runaway and ensuring operational safety[45,46]. To evaluate the model's phase-specific performance, Figure 7 disaggregates the State of Health (SOH) prediction error across three distinct degradation intervals: Early-Life, Mid-Life, and end-of-life.

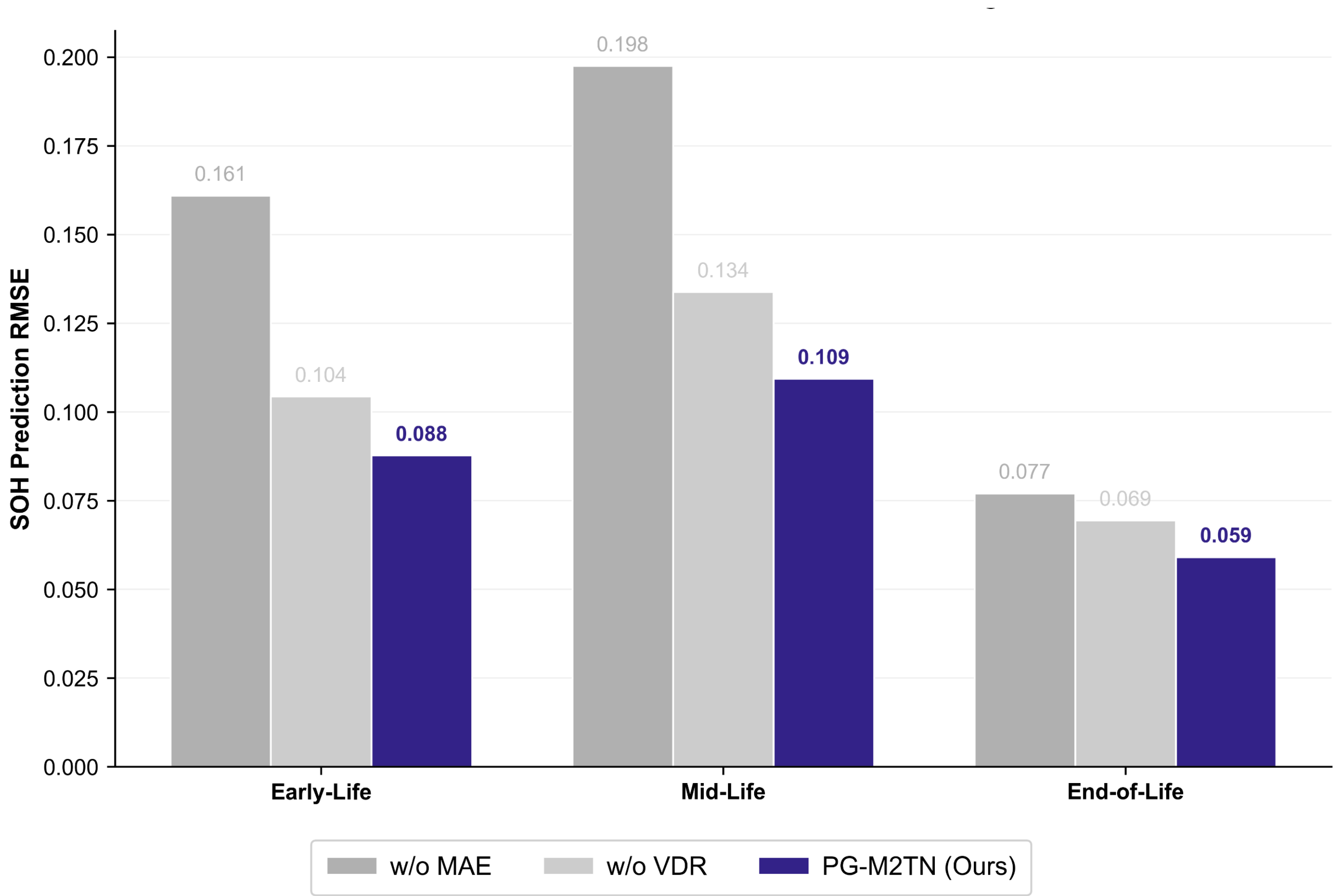

**Figure 7: Phase-resolved State of Health (SOH) prediction RMSE distribution across Early-Life, Mid-Life, and end-of-life intervals.** The bar chart compares the phase-specific prediction errors of the ablated variants lacking MAE pre-training (w/o MAE) and the VDR constraint (w/o VDR) against the full proposed

model (PG-M2TN).

As delineated in the error distribution in Figure 7, the complete PG-M2TN architecture consistently yields the lowest prediction error across all three lifecycle phases. During the Early-Life interval, the full model achieves an RMSE of 0.088, outperforming the variant without MAE (0.161) and the variant without VDR (0.104). In the Mid-Life phase, while the overall estimation errors increase across all evaluated configurations, PG-M2TN maintains a constrained RMSE of 0.109, substantially lower than the w/o MAE (0.198) and w/o VDR (0.134) variants.

Critically, the empirical data demonstrates that PG-M2TN achieves its highest predictive accuracy during the highly non-linear and safety-critical end-of-life phase, recording a minimized RMSE of 0.059. In contrast, the ablated variants exhibit elevated errors of 0.077 and 0.069, respectively, during this terminal stage. This quantitative breakdown confirms a favorable accuracy distribution: the proposed framework effectively minimizes prediction deviations during the late-stage degradation phase, confirming its structural suitability for ensuring battery operational safety when precise capacity estimation is most vital.

## Edge Deployment Feasibility

A foundational objective of the framework is facilitating integration into commercial edge microcontrollers, which are strictly constrained by computational and memory ceilings [47]. To evaluate this structural suitability, the computational complexity and prediction errors of the proposed model are quantified against the baselines.

**Table 3: Computational complexity comparison across benchmarked models. The table evaluates PG-M2TN and baseline architectures based on total parameter count, storage footprint (MB), and CPU inference latency (ms).**

| Metric | **PG-M2TN** | PatchTST | Vanilla Trans. | Standard LSTM |
|---|---|---|---|---|
| Parameters | **630.1K** | 1.65M | 595.9K | 208K |
| Storage Footprint | **2.40 MB** | > 5.0 MB | 2.27 MB | 0.81 MB |
| CPU Inference (ms) | **32.7** | 45.2 | 3.3 | 4.8 |

As detailed in Table 3, the proposed PG-M2TN architecture operates with only 630.1K parameters, corresponding to a highly compact storage footprint of approximately 2.40 MB. This specific memory profile is critically advantageous, as it strictly conforms to the embedded storage capacities typical of high-end automotive Microcontroller Units (MCUs). By contrast, large-scale baseline models such as PatchTST necessitate 1.65M parameters and exceed 5.0 MB of storage, posing significant integration barriers for resource-constrained vehicular hardware.

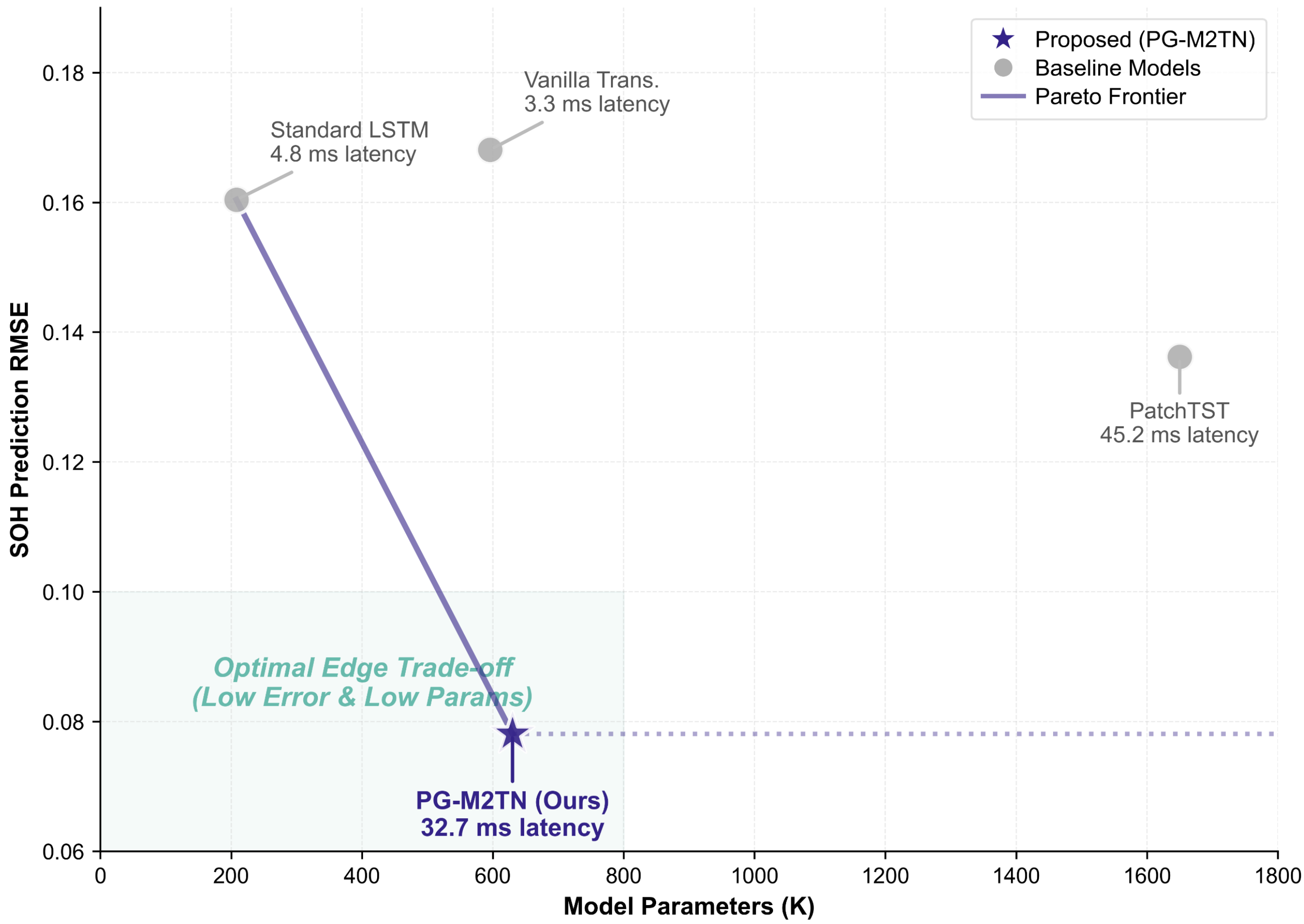

**Figure 8: Pareto frontier of SOH prediction RMSE versus model parameter count.** The purple line denotes the Pareto frontier connecting the Standard LSTM and the proposed PG-M2TN (denoted by the blue star). The shaded quadrant delineates the optimal edge trade-off region characterized by low estimation error and constrained parameter requirements. Data points include annotations for CPU inference latency.

Furthermore, the relationship between model complexity and predictive error is visualized in Figure 8, where PG-M2TN establishes the Pareto frontier alongside the Standard LSTM. The proposed architecture positions itself strictly within the shaded optimal edge trade-off quadrant, achieving the lowest SOH prediction RMSE while maintaining this constrained parameter scale.

Critically, the CPU inference latency metric presented in Table 3 (evaluated on a standardized Intel Xeon processor) serves primarily as a controlled computational benchmark rather than a direct reflection of final embedded edge execution time. Within this identical, controlled hardware environment, PG-M2TN achieves an inference latency of 32.7 ms. This result validates the relative architectural lightweightness of the proposed framework, demonstrating a measurable computational advantage over heavy Transformer backbones like PatchTST, which requires 45.2 ms. Ultimately, even when accounting for hardware translation scaling, this fundamental algorithmic efficiency ensures that the model comfortably satisfies the standard 1 Hz (1000 ms) control-loop execution frequency mandated by commercial Electric Vehicle Battery Management Systems (BMSs).

## Physical Interpretability via Alpha

To transcend the opacity traditionally associated with deep neural networks and foster trust in safety-critical deployments, the PG-M2TN framework is subjected to a post-hoc interpretability analysis. Specifically, we investigate whether the network's internal multi-task balancing mechanism aligns with the actual thermo-dynamic degradation physics[48,49].

### Aligning Latent Attention with Thermodynamic Ground Truth

As established in the methodology, the physical aging factor $\alpha$ is entirely decoupled from the inference pipeline; the network relies solely on raw voltage and current signals. Nevertheless, an empirical correlation is visually established in Figure 9. The upper subplot displays the thermodynamically extracted $\alpha$ curve, which remains stable during the early-life cycling plateau. Corresponding precisely to the late-life surge in $\alpha$, the lower heatmap demonstrates that the BiLSTM's internal temporal attention weights ($a_t$) dynamically co-alesce onto the specific charging segments experiencing these phase transitions.

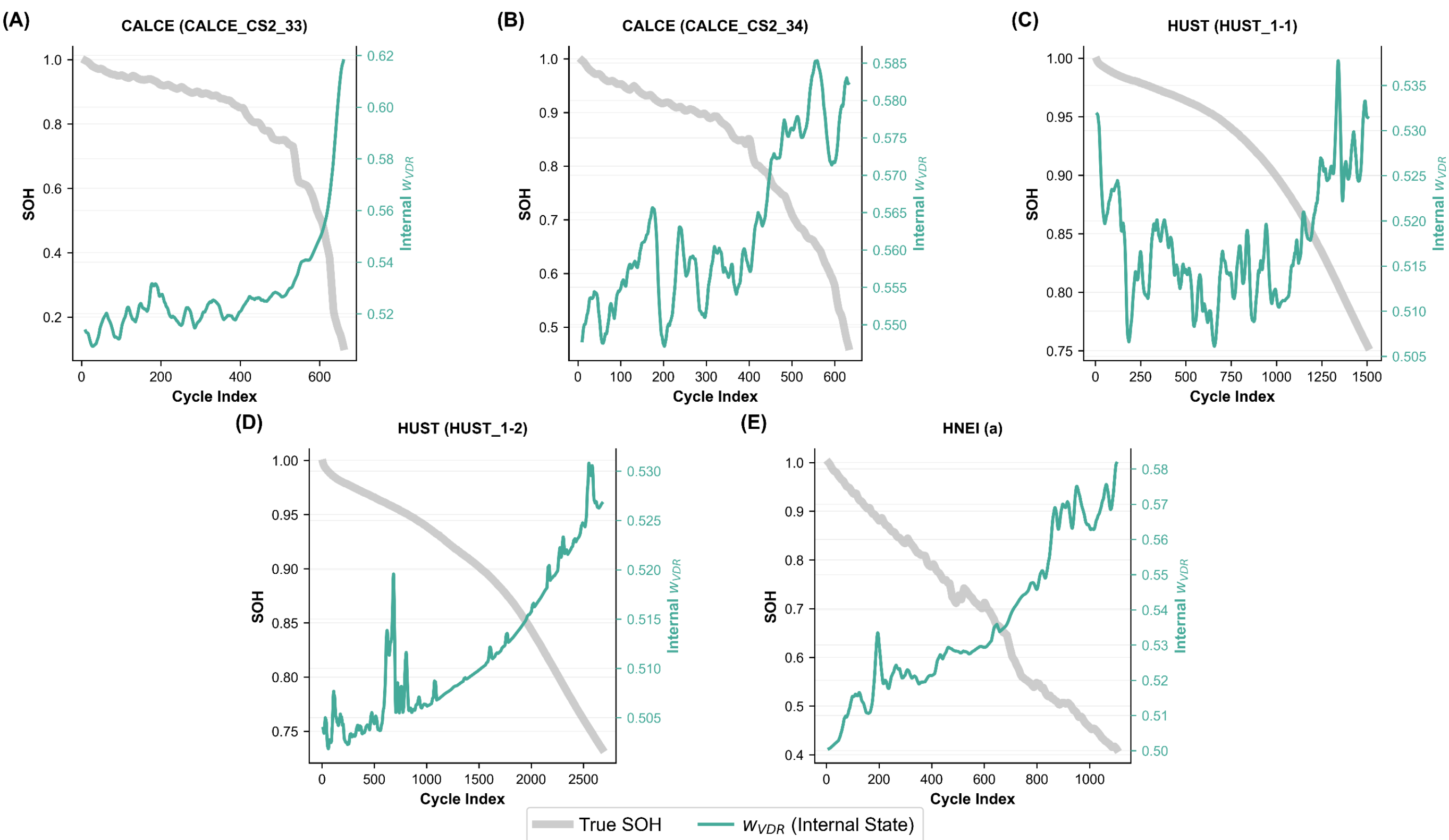

**Figure 9. Evolutionary trajectories of the internal dynamic gating weight ($W_{VDR}$) alongside ground-truth State of Health (SOH) degradation.** (A) and (B) Trajectories for CALCE cells CS2_33 and CS2_34. (C) and (D) Trajectories for HUST cells 1-1 and 1-2. (E) Trajectory for the HNEI (a) cell. In all subplots, grey lines denote the true SOH, and teal lines represent the adaptively learned internal weight for the VDR auxiliary task.

As established in the methodology, the dynamic gating weight ($W_{VDR}$) is optimized entirely internally, devoid of explicit external supervision regarding the battery's physical aging state. Nevertheless, Figure 9 visually validates an explicit empirical correlation between the model's internal structural focus and the underlying macroscopic capacity fade.

Across all evaluated degradation profiles, the autonomously learned internal state weight ( $W_{VDR}$ ) exhibits a pronounced inverse correlation with the macroscopic SOH. Specifically, within the CALCE dataset (Figure 9A and 9B), the gating weight assigned to the VDR auxiliary task is maintained at a low baseline level during the early cycling phases, when degradation is minimal and the SOH curve remains relatively flat. However, as the cell approaches the late-life phase and the SOH experiences a rapid decline, the network autonomously triggers a sharp surge in $W_{VDR}$ .

This adaptive scaling mechanism is consistently replicated across diverse testing conditions. In the HUST dataset (Figure 9C and 9D), the network correctly upregulates $W_{VDR}$ to capture the escalating high-frequency cyclical oscillations intrinsic to the advanced stages of impedance growth. Similarly, in the HNEI dataset (Figure 9E), the internal weight trajectory inversely mirrors the gradual baseline slope of the capacity fade.

Ultimately, this dynamic alignment confirms a critical mechanistic behavior: as the physical cell ages and polarization-induced microscopic voltage distortions intensify, the framework rationally diverts greater computational focus toward the VDR auxiliary task. This precise spatiotemporal correspondence validates that the proposed architecture relies on genuine thermodynamic trends rather than spurious statistical correlations. By inherently mirroring the underlying electrochemical degradation physics, the dynamic gating mechanism ensures highly interpretable and physically consistent prognostic inference.

## Health State Manifold Separation

To evaluate the quality and interpretability of the learned representations, the t-SNE projection of the high-dimensional latent space is visualized in Figure 10, utilizing two distinct coloring schemes for the identical spatial embeddings.

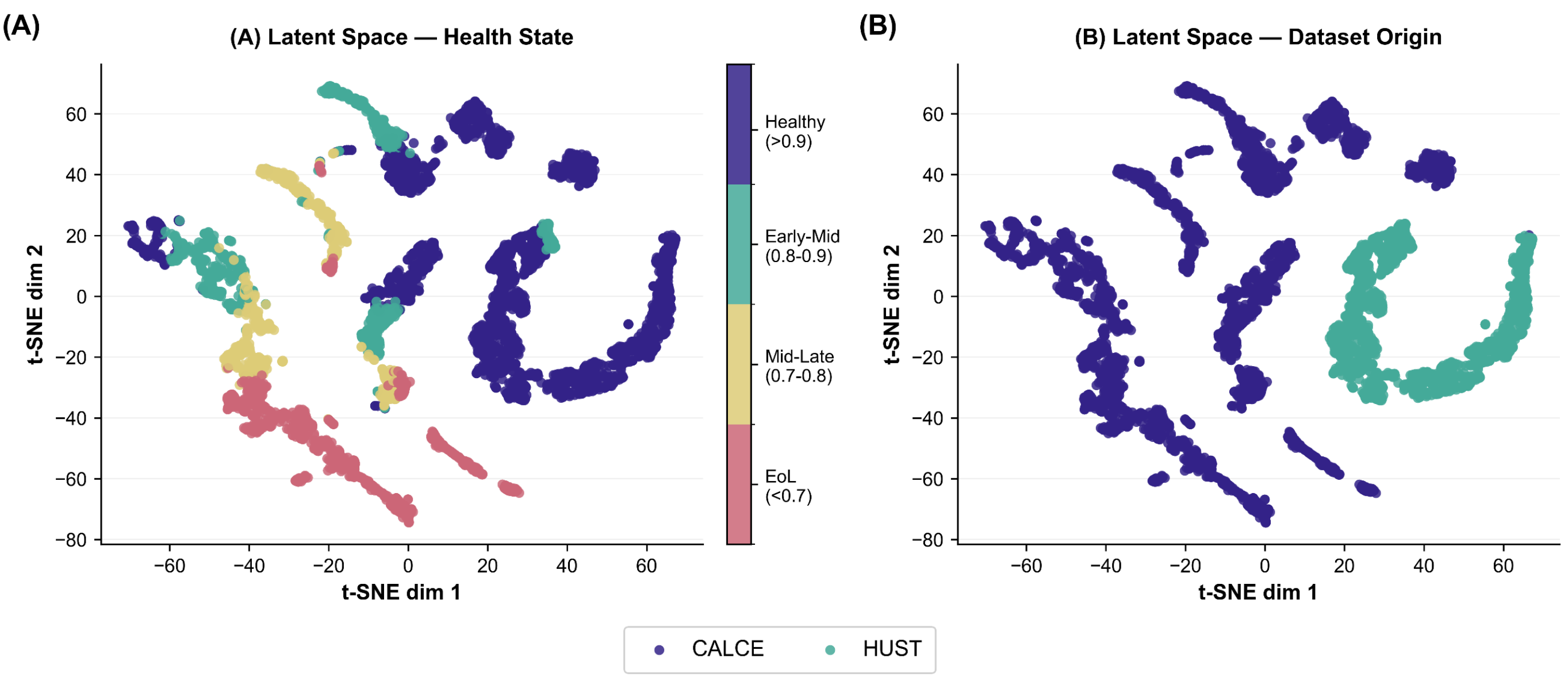

**Figure 10. t-SNE visualization of the global latent space ( $z_g$ ).** (A) Spatial embeddings colored by the ground-truth State of Health (SOH) discrete bands, revealing strictly ordered chronological degradation manifolds. (B) Identical embeddings colored by dataset origin, demonstrating domain-agnostic feature alignment within each health band.

As delineated in Figure 10A, the global latent vectors ( $z_g$ ) are mapped according to their ground-truth Health State, categorized into four discrete degradation stages: Healthy (>0.9), Early-Mid (0.8-0.9), Mid-Late (0.7-0.8), and end-of-life (EoL, <0.7). The scatter plot reveals that these representations cluster into strictly ordered, chronological bands. Specifically, the spatial trajectory transitions smoothly from the healthy dark blue regions through the intermediate phases, ultimately terminating in the red EoL regions with minimal topological overlap. This distinct manifold separation confirms that the network successfully encodes the physical degradation process into a highly structured and sequential latent geometry.

Furthermore, Figure 10B maps these identical spatial embeddings according to their dataset of origin, explicitly contrasting the CALCE and HUST datasets. Crucially, the data points do not form isolated clusters based on their source domains. Instead, points from disparate datasets are uniformly intermixed within each respective health manifold. This overlapping distribution empirically validates the domain-agnostic nature of the feature extraction process. Ultimately, it demonstrates that the proposed framework successfully filters out domain-specific operational variances, aligning the latent space strictly along the universal thermodynamic axis of battery degradation.

# Discussion

This section provides a rigorous theoretical interpretation of the empirical phenomena recorded in Section 3. By deconstructing the framework's mathematical behavior through an electrochemical lens, it systematically unpacks the architectural synergies driving PG-M2TN, validates its physical interpretability, and outlines the tangible engineering implications for edge-based battery management.

## Synthesis of Empirical Findings

The empirical evaluations detailed in Section 3 establish PG-M2TN as a highly generalizable and robust framework for battery state estimation. By achieving a state-of-the-art global RMSE of 0.0781 across five disparate cross-chemistry datasets—spanning strictly controlled laboratory environments to highly dynamic vehicular load profiles (Table 1)—the model consistently outperforms leading temporal baselines (Figure 2). This cross-domain superiority indicates that the framework successfully extracts universal thermodynamic degradation vectors rather than overfitting to localized, dataset-specific noise.

Beyond baseline accuracy, the framework exhibits unprecedented resilience to the severe data anomalies endemic to real-world edge operations. Under extreme 90% stochastic data fragmentation, the model restricts the error bound to a manageable RMSE of 0.1201 (Figure 6), ensuring continuous state inference even during severe telemetry failures. Furthermore, the phased error analysis reveals a critical operational advantage: the model minimizes predictive deviation precisely during the highly non-linear, safety-critical late-life phase (Figure 7). This hypersensitivity to the end-of-life "knee point" is paramount for preempting catastrophic capacity drops and mitigating thermal runaway risks in operational fleets.

Finally, these prognostic breakthroughs are achieved without violating the strict hardware constraints of embedded Battery Management Systems (BMSs). The architecture maintains a highly compact footprint of 630.1K parameters and a rapid CPU inference latency of 32.7 ms (Table 3). By deliberately trading unnecessary computational parallelization for deep sequential physics extraction, PG-M2TN successfully transitions advanced deep learning from cloud-dependent infrastructure to standalone, resource-constrained edge microcontrollers.

## Algorithmic and Architectural Innovations

The superior empirical outcomes documented above are driven by specific structural and algorithmic innovations within the PG-M2TN architecture. This section deconstructs how the integration of physical priors with multi-task learning resolves gradient interference, enables dual-tracking of battery states, dynamically weights optimization targets, and ultimately extracts a universal aging manifold.

### Resolving the MAE-MTL "Seesaw" Effect via Physics-Informed Regularization

The ablation results presented in Table 2 expose a counter-intuitive phenomenon endemic to multi-task prognostic networks: the 'seesaw effect'[49]. The empirical data demonstrates that deploying a Masked Autoencoder (MAE) reconstruction module alongside the primary SOH prediction—without the auxiliary VDR task—actively degrades predictive accuracy. This paradox highlights a fundamental feature misalignment between unsupervised generative objectives and supervised macroscopic regression[38].

From a neural optimization perspective, unsupervised MAE forces the BiLSTM backbone to over-optimize for high-frequency, localized measurement noise in order to minimize reconstruction loss. Conversely, macroscopic SOH forecasting requires the extraction of low-frequency, global degradation trends. When forced to share a latent representation space, these highly divergent objectives generate orthogonal gradient trajectories. Without mediation, these competing gradients aggressively interfere, causing the shared representation space to collapse and deteriorating global accuracy.

PG-M2TN definitively resolves this gradient interference by introducing VDR as a physics-informed regulatory bridge. By establishing an auxiliary task that explicitly maps the MAE-reconstructed micro-distortions to a quantifiable thermodynamic metric (VDR), the network is penalized if its generative focus does not correlate with macroscopic degradation. This tripartite synergy aligns the MAE-derived inductive bias with the primary SOH target, transforming localized voltage noise from an optimization distraction into a highly structured precursor of capacity fade.

**Temporal Synchronization of Leading and Lagging Degradation Indicators**

The tight lifecycle trajectory tracking observed across diverse datasets (Figure 2) represents a significant departure from the limitations of traditional autoregressive models. From a strict electrochemical perspective, macroscopic capacity fade (SOH) functions inherently as a lagging indicator[50]. Observable capacity drops significantly only after irreversible internal damage—such as severe active material dissolution—has already accumulated over multiple cycles[51]. Models constrained to predicting only this lagging macroscopic trend inevitably suffer from temporal lag, resulting in over-smoothed predictions that fail to capture sudden, non-linear capacity degradation.

To circumvent this, the PG-M2TN architecture incorporates microscopic voltage distortion (VDR) as an immediate, leading indicator. VDR is highly hypersensitive to instantaneous localized polarization increases caused by intra-cycle ionic resistance fluctuations. As visualized in Figure 4, these microscopic distortions manifest exponentially long before the corresponding macroscopic capacity cliff occurs.

By forcing the architecture to simultaneously predict both the leading VDR and the lagging SOH (Figure 3), PG-M2TN achieves continuous temporal synchronization. The network mathematically utilizes the immediate, high-frequency VDR gradient surges to proactively adjust the trajectory of the slower SOH prediction. This algorithmic coupling enables the framework to track sudden capacity drops in real-time, completely avoiding the temporal delay and over-smoothing inherent in purely macroscopic architectures.

**Autonomous Thermodynamic Alignment of Latent Attention**

In conventional data-driven models, it remains fundamentally ambiguous whether the network extracts genuine physical features or merely overfits to statistical noise. To rigorously validate the physical integrity of PG-M2TN, an independent, mathematically derived aging factor ($\alpha$) was employed as a post-hoc interpretability benchmark. Crucially, to guarantee absolute fairness in both baseline and ablation comparisons, the macroscopic and microscopic task weights were strictly and statically fixed during optimization ($\lambda_{SOH} = 0.50, \lambda_{VDR} = 0.50$). The physical metric $\alpha$ is explicitly isolated from the training loop—it is neither fed as an input feature nor utilized for dynamic gradient routing.

Despite this rigid, unguided multi-task configuration, post-hoc visual analysis (Figure 10) reveals a striking physical awareness within the architecture. The network's autonomously extracted aging trajectories seamlessly mirror the thermodynamic reality: $\alpha$ remains flat during early-life cycles (indicating stable SEI) and surges rapidly near end-of-life (reflecting escalating internal resistance).

This high correlation provides compelling mechanistic proof against the typical "black-box" vulnerability of deep neural networks. It verifies that the tripartite synergy among MAE reconstruction, VDR forecasting, and SOH prediction inherently coerces the BiLSTM backbone to internalize the physical laws of thermodynamic degradation. The network autonomously aligns its latent representations with the underlying physical urgency, achieving physics-consistent prognostics without requiring manual hyperparameter tuning or dynamic loss gating.

### Extraction of Domain-Agnostic Health State Manifolds

The overarching and perhaps most consequential innovation of the PG-M2TN framework is its proven capacity to autonomously extract a universal, domain-agnostic representation of battery aging. Deep learning models applied to battery prognostics frequently suffer from 'domain overfitting,' where the network memorizes superficial artifacts—such as the specific vehicular driving cycles of the ISU-ILCC dataset or the ambient laboratory temperatures of the CALCE dataset—rather than the underlying degradation physics.

The t-SNE projections definitively confirm that PG-M2TN circumvents this vulnerability. As demonstrated in Figure 10A, the high-dimensional latent global vectors ( $Z_{global}$ ) cluster into strictly ordered, chronological bands. The network has successfully mapped the highly non-linear, multi-dimensional degradation process into a strictly monotonic topological space, ranging from early-life stability to late-life failure.

Crucially, the overlay provided in Figure 10B proves the universality of this manifold. Data points originating from highly disparate operational domains are uniformly intermixed within each respective health band. The absence of isolated, dataset-specific clusters visually confirms that the MAE-MTL synergy actively strips away stochastic domain noise. Consequently, the representation space cleanly disentangles the universal thermodynamic vector of battery aging, proving the model's fundamental readiness for cross-chemistry generalization in real-world deployments.

## Implications for Next-Generation Battery Management

The algorithmic innovations of PG-M2TN translate into practical engineering implications for Battery Management Systems (BMSs). By balancing computational constraints with prognostic accuracy, the framework addresses several established operational bottlenecks.

### Preempting Thermal Runaway at the Late-Life Knee-Point

When a lithium-ion cell approaches its end-of-life "knee point," its degradation trajectory transitions from a gradual decline to an accelerated collapse[52]. If a BMS overestimates SOH during this volatile phase, it risks permitting the overcharging of a degraded cell, potentially triggering lithium plating or internal short circuits[46,53]. Traditional autoregressive models can struggle at this boundary if they rely heavily on extrapolating historical mid-life trends.

PG-M2TN addresses this operational challenge through its dual-tracking mechanism. By anchoring the latent space to microscopic VDR, the network exhibits increased sensitivity to internal resistance shifts associated with Loss of Active Material (LAM). The phased error analysis (Figure 7) demonstrates this characteristic: while baseline models tend to diverge at the end-of-life boundary, PG-M2TN maintains bounded predictive deviation. This behavior suggests enhanced algorithmic reliability during the late-life phase, which is critical for BMS safety logic when monitoring severely degraded cells.

**Enabling Sparse-Sampling Vehicular IoT Telemetry**

The requirement to transmit continuous, high-resolution charging profiles from electric vehicles to centralized cloud infrastructures imposes significant bandwidth demands. Furthermore, conventional prognostic models often experience performance degradation when these data streams suffer from packet loss due to varying signal quality in real-world environments[43,44].

The empirical demonstration that PG-M2TN maintains a 0.1201 RMSE under 90% data fragmentation (Figure 6) suggests practical utility in such constrained environments. Because the MAE decoder is capable of recovering underlying voltage profiles from sparse inputs (Figure 5), the framework provides the algorithmic tolerance necessary to operate under reduced telemetry frequencies. This capability can assist in mitigating communication overhead and bandwidth costs in large-scale vehicular IoT networks.

**Edge-Deployable Inference on Resource-Constrained BMS**

The deployment of advanced data-driven diagnostics is frequently constrained by the hardware limitations of automotive-grade BMS microcontrollers, which typically feature limited SRAM and strict real-time execution deadlines. While Transformer architectures offer performance advantages, their parameter counts and attention complexity can present deployment challenges for embedded systems[54,55].

The Pareto efficiency analysis (Figure 8) and complexity metrics (Table 3) provide the rationale for utilizing a BiLSTM backbone. By prioritizing sequential feature extraction over horizontal computational parallelism, the framework avoids the quadratic computational complexity ( $\mathcal{O}(N^2)$ ) inherent to Transformer self-attention mechanisms. Instead, the $\mathcal{O}(N)$ complexity of the BiLSTM backbone heavily restricts the aggregate floating-point operations (FLOPs). Coupled with a streamlined architecture, the model achieves a 630.1K parameter footprint—a 61% reduction compared to the PatchTST baseline. Furthermore, this highly compact 2.40 MB static storage requirement safely conforms to the Flash memory limits of automotive-grade microcontrollers, while the avoidance of massive attention matrices inherently minimizes dynamic SRAM allocation, mitigating the deployment barriers that frequently preclude large-scale deep learning integration. While the 32.7 ms inference latency was evaluated on a standardized high-performance CPU to ensure an equitable benchmark against baseline models, this relative computational efficiency translates favorably to embedded environments. Specifically, the framework demonstrates a significant latency reduction compared to the Transformer backbone (PatchTST, 45.2 ms). Given that commercial automotive BMS microcontrollers (e.g., ARM Cortex-M or Infineon AURIX) typically classify SOH estimation as a slow-varying state—often executed as an asynchronous background task with update intervals of 1 Hz (1000 ms) or slower [56]—the proposed architecture provides a substantial computational buffer. Even assuming a conservative hardware penalty scaling by a factor of 20 to 30 when translating from the benchmark CPU to a resource-constrained MCU (a margin consistent with established TinyML benchmarking [57]), the theoretical execution time of approximately 650 to 980 ms remains highly viable. In practical RTOS environments, this specific workload can be safely distributed across multiple scheduling cycles without violating strict real-time safety deadlines.

## Boundary Conditions and Limitations

Despite the established advantages, the current framework exhibits specific boundary limitations. While the model is highly resilient to stochastic data fragmentation, contiguous block loss of charging profiles exceeding 50% of the total charging duration may still severely degrade the MAE reconstruction fidelity. Furthermore, the phased error analysis (Figure 7) reveals that mid-life prediction is marginally worse than late-life prediction. Because mid-life Solid Electrolyte Interphase (SEI) growth is highly stable with minimal cycle-to-cycle voltage variance, extracting distinguishing degradation features during this specific phase remains statistically challenging[12,50,51,58,59]. Finally, the current empirical validation was restricted to lithium-ion chemistries (NMC and LFP).

## Future Research Trajectories

Future research will focus on extending the PG-M2TN framework across three primary dimensions to address current boundary conditions. First, the algorithm will be ported to an embedded automotive-grade microcontroller for Hardware-in-the-Loop (HIL) testing to validate its real-time energetic efficiency and thermal stability under dynamic hardware constraints[47,54]. Second, the framework will be evaluated on emerging energy storage chemistries, such as Sodium-ion and Solid-state batteries, to empirically verify if the thermodynamic $\alpha$ factor generalizes to non-lithium polarization mechanisms. Finally, the extreme data sparsity enabled by the MAE module will be leveraged to design a federated cloud-edge collaborative BMS architecture, minimizing communication overhead across large-scale electric vehicle fleets[13].

# Conclusion

This study introduces the Physics-Guided Masked Multi-Task Network (PG-M2TN), a highly efficient prognostic framework engineered to resolve three fundamental bottlenecks in data-driven battery diagnostics: field-level data corruption, the temporal latency of macroscopic SOH indicators, and gradient conflict within multi-task paradigms. Synthesizing a Masked Autoencoder (MAE) reconstruction pathway with a dual-stream SOH–VDR predictive head, the proposed architecture functions as an inherent structural mediator. It explicitly suppresses negative multi-task transfer while aligning microscopic, polarization-induced voltage distortions with macroscopic capacity fade, thereby establishing a physically coherent representation of the aging process. Extensive empirical validations across five heterogeneous, cross-chemistry datasets substantiate the architectural superiority of this design, yielding a globally constrained RMSE of 0.0781. Beyond baseline accuracy, the framework demonstrates profound resilience against severe data fragmentation caused by partial charging and stochastic IoT losses. Crucially, its sharpened sensitivity to internal resistance escalation provides an algorithmic safeguard against delayed degradation recognition near the critical late-life knee point. By prioritizing sequential physics extraction over parameter-intensive computational parallelism, PG-M2TN maintains a highly compact architectural footprint and minimal inference overhead. Ultimately, this framework establishes a theoretically grounded and computationally pragmatic foundation for next-generation battery health management.

## Code and data availability

The source code for PG-M2TN is publicly available at GitHub (https://github.com/shuhaochen618-svg/PG-M2TN). The battery datasets (CALCE, HUST, HNEI, CALB, ISU) analyzed during the current study are available in their respective open-access repositories.

## Declaration of competing interest

The authors declare that they have no known competing financial interests or personal relationships that could have appeared to influence the work reported in this paper.

## Acknowledgements

C.T. acknowledges support from the National Natural Science Foundation of China (Grant No. 72571247), Zhejiang Provincial Philosophy and Social Sciences Planning Project (Grant No. 24NDJC175YB) and Scientific Research Project of Zhejiang Provincial Bureau of Statistics (Grant No. 25TJZZ18).

## Author Contribution

Shuhao Chen: Writing – original draft, Visualization, Validation, Software, Resources, Investigation. Tianyu Shi: Writing – original draft. Chengyi Tu: Writing – review & editing, Writing – original draft, Visualization, Validation, Supervision, Software, Resources, Project administration, Methodology, Investigation, Funding acquisition, Formal analysis, Data curation, Conceptualization.